\documentclass[letterpaper, 10 pt, conference]{ieeeconf}  

\IEEEoverridecommandlockouts                              

\usepackage{cite}
\usepackage{amsmath,amssymb,amsfonts}
\usepackage{algorithm} 
\usepackage{algpseudocode} 
\usepackage{graphicx}
\usepackage{textcomp}
\usepackage{xcolor}
\usepackage{hyperref}
\usepackage{caption,subcaption}
\usepackage{multirow}

\title{\LARGE \bf
LIDAR data based Segmentation and Localization using Open Street Maps for Rural Roads
}

\author{Stephen Ninan$^{1}$ and Sivakumar Rathinam$^{2}$
\thanks{*This work was funded by a grant from the U.S. Department of Transportation, University Transportation Centers Program.}
\thanks{$^{1}$ Graduate Student, Texas A\&M University, College Station, TX 77843, USA
        {\tt\small nstephen@tamu.edu}}%
\thanks{$^{2}$ Faculty Member, Texas A\&M University, College Station, TX 77843, USA
        {\tt\small srathinam@tamu.edu}}%
}

\begin{document}

\maketitle
\pagestyle{plain}
\setcounter{page}{1}

\begin{abstract}
Accurate pose estimation is a fundamental ability that all mobile robots must posses in order to traverse robustly in a given environment. Much like a human, this ability is dependent on the robot's understanding of a given scene. For Autonomous Vehicles (AV's), detailed 3D maps created beforehand are widely used to augment the perceptive abilities and estimate pose based on current sensor measurements. This approach however is less suited for rural communities that are sparsely connected and cover large areas. To deal with the challenge of localizing a vehicle in a rural setting, this paper presents a data-set of rural road scenes, along with an approach for fast segmentation of roads using LIDAR point clouds. The segmented point cloud in concert with road network information from Open Street Maps (OSM) is used for pose estimation. We propose two measurement models which are compared with state of the art methods for localization on OSM for tracking as well as global localization. The results show that the proposed algorithm is able to estimate pose within a 2 sq. km area with mean accuracy of 6.5 meters.

\end{abstract}

\section{INTRODUCTION}

Perception and localization technologies play critical roles in the operations of mobile robots and  Autonomous Vehicles (AV's). In general, a vehicle's ability to safely navigate a given environment is dependent on its understanding of the environment, and its pose within the environment. Identification of a drivable\footnote{Part of the road surface that the vehicle can drive on.} surface, and the localization of the vehicle within this surface are therefore two fundamental tasks that an AV needs to perform in real time.\par
Road segmentation, which is part of the drivable surface identification, has been extensively researched in the literature, and a number of algorithms have been proposed for this task. Similarly, a large number of data-sets and corresponding benchmarks have been established, e.g. the KITTI road data-set \cite{6728473} provides annotated road images along with LIDAR data. More recent data-sets such as CaRINA \cite{7795529} provide road detection benchmarks from selected Brazilian urban scenarios, NuScenes \cite{9156412} provides data from around Boston and Singapore, Cityscapes \cite{Cordts2016Cityscapes} contains semantic data from 50 cities in and around Germany.\par
A common feature among these data-sets is that they contain very little to no data from a rural setting. In contrast to an urban driving environment, roads in rural communities may not contain key features such as sidewalks, curbs, lane markings which characterize the boundaries of roads in urban communities. As a result, the segmentation algorithms based on existing data-sets may not perform well in never before seen environments involving rural communities. 

\par
The most common approach \cite{levinson2010robust} to vehicle localization involves creating a dense map of the environment, by mapping the environment before hand. This map is then used to estimate the vehicle's pose based on sensor readings and inertial measurements in a Monte Carlo scheme. The drawback here being that any route to be traversed, must be mapped beforehand. Another approach to localization is to replace dense maps with freely available topological maps that encode information about road shapes. The road network information is then combined with the information about the surrounding environment retrieved from sensors.

\par
The contributions of our work are three fold: First, a newly annotated rural road data-set has been developed and presented in this paper. To our knowledge, this is the first data-set which contains annotated road labels for a variety of rural terrains. We believe this will be a first step towards the implementation of AV's for rural communities. Second, we make use of the developed data-set to propose a Range Image based road segmentation approach for LIDAR point clouds. Third, we utilize the trained road segmentation model in conjunction with a Particle Filter for GPS denied pose estimation with the use of topological map networks, for tracking (initial pose known) as well as global localization (initial pose unknown). 

\par
The rest of this paper is organized as follows: Section \ref{prior_work} contains a summary of the existing work on road segmentation and localization using topological maps. In Section \ref{method}, we explain the underlying framework for road segmentation and localization. Section \ref{data-set} introduces the rural road data-set, followed by a discussion on the results from model training and pose estimation tests, in Section \ref{train}.

\section{Related Work} \label{prior_work}
A survey of the point cloud segmentation algorithms suggests two main approaches - a feature based approach and a learning based approach. The first approach  proposes using features such as point elevation, change in point height with respect to neighbouring points and distance to nearest points to characterize a point. A point is then classified as a road point if its feature values lie within a predetermined threshold. The algorithm proposed by Zhang et al. \cite{8291612} is an example of this approach.\par

The second class is a broad family of Deep Learning approaches.  3DShapenet\cite{7298801} and Voxnet \cite{7353481} are examples of volumetric CNN's that make use of 3D convolutions for segmentation. The drawback though, is that 3D convolutions are computationally expensive, increasing the run-time of these algorithms. In PointNet \cite{qi2017pointnet}, the authors propose a network to directly deal with point clouds. Lastly, point cloud projection to either Bird's Eye View (BEV) or Spherical Front View (SFV) has been successfully used to reduce point clouds to 2D images. In \cite{caltagirone2017fast}, Caltagiron et al. propose a top view road detection approach using a Fully Convolutional Network (FCN). Similarly, SalsaNet \cite{aksoy2020salsanet} employs a BEV projection on KITTI data-set for road and vehicle detection. Likewise, front view or perspective projections have been used in Squuezeseg \cite{wu2018squeezeseg} and Multi-view CNN \cite{qi2016volumetric}. 

\par
The use of topological maps for localization have been widely researched \cite{hentschel2010autonomous} as an alternative to traditional mapping. Most of the existing literature \cite{8870918} relies on a Particle Filter for pose estimation, however the differences between them is in the measurement models they employ. In OpenStreetSLAM \cite{floros2013openstreetslam}, the authors propose a chamfer matching algorithm to estimate a vehicles pose using odometry measurements. Chen et al. \cite{zhou2021efficient} learn a model to embed road images to their corresponding map tiles, which is then used as a measurement model. Ruchti et al. \cite{ruchti2015localization} employ a zero mean Gaussian as an observational model and in MapLite \cite{ort2019maplite}, the authors propose a signed distance function for the the measurement model with good results.

\begin{figure}[b]
    \centering
    \includegraphics[width=0.45\textwidth]{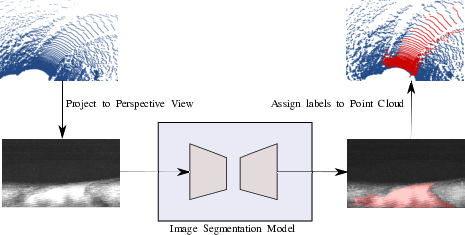}
    \caption{Input and Output for proposed model}
    \label{fig:galaxy}
\end{figure}
\section{Methodology} \label{method}
\subsection{Road Segmentation}

Existing literature suggests that simple heuristic based approaches are inadequate for accurate road segmentation in a rural scenario. The large amount of variability within what is deemed as a drivable surface in a rural scene, means that limiting the algorithms to a finite feature space is not a suitable approach. 
A schematic of our proposed segmentation pipeline is shown in Figure 1. Rather than suggesting a new model tailored for rural roads, we propose projecting the point clouds to a Spherical Front View (SFV) Image, which is then passed to am image segmentation model. The segmentation result for each pixel is then mapped to the corresponding point in the point cloud. In contrast to the previous work, our proposed approach aims to leverage the ability of CNN's to learn low level as well as high level image features. This approach eliminates the need for careful feature selection followed by tuning of the feature threshold values. \par

\begin{figure*}[h]
\centering
\begin{subfigure}{0.23\textwidth}
\centering
\includegraphics[width=\textwidth]{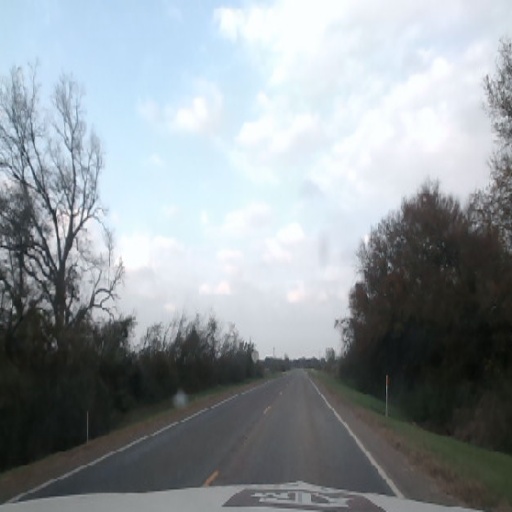}
\end{subfigure}  
\vspace{0.1cm}
\begin{subfigure}{0.23\textwidth}
\centering
\includegraphics[width=\textwidth]{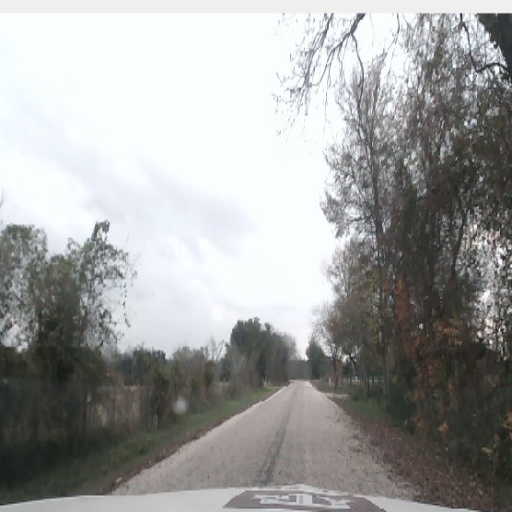}
\end{subfigure}
\begin{subfigure}{0.23\textwidth}
\centering
\includegraphics[width=\textwidth]{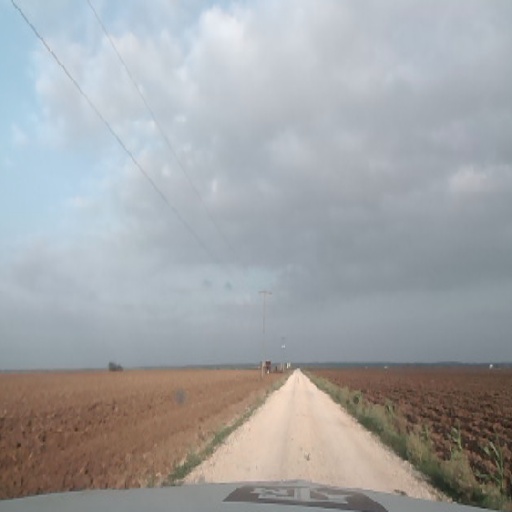}
\end{subfigure}
\begin{subfigure}{0.23\textwidth}
\centering
\includegraphics[width=\textwidth]{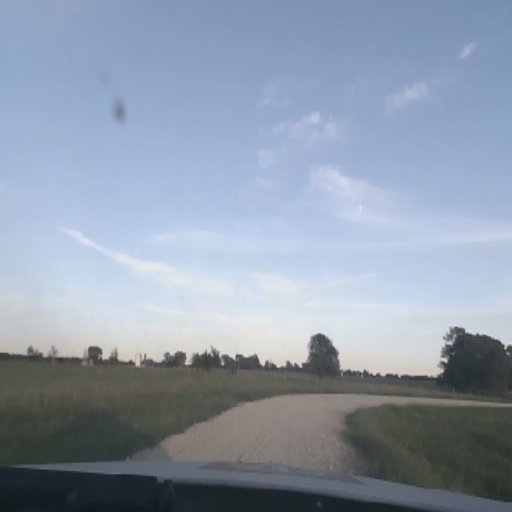}
\end{subfigure} 
\vspace{0.1cm}
\begin{subfigure}{0.23\textwidth}
\centering
\includegraphics[width=\textwidth]{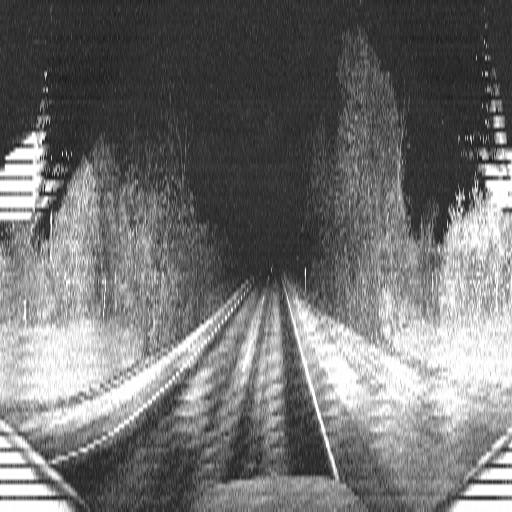}
\end{subfigure}
\begin{subfigure}{0.23\textwidth}
\centering
\includegraphics[width=\textwidth]{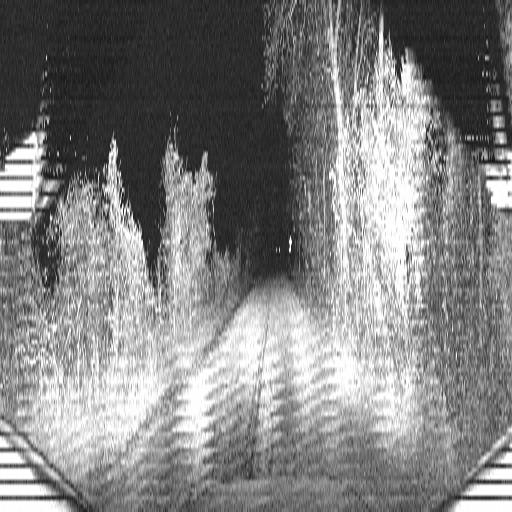}
\end{subfigure}
\begin{subfigure}{0.23\textwidth}
\centering
\includegraphics[width=\textwidth]{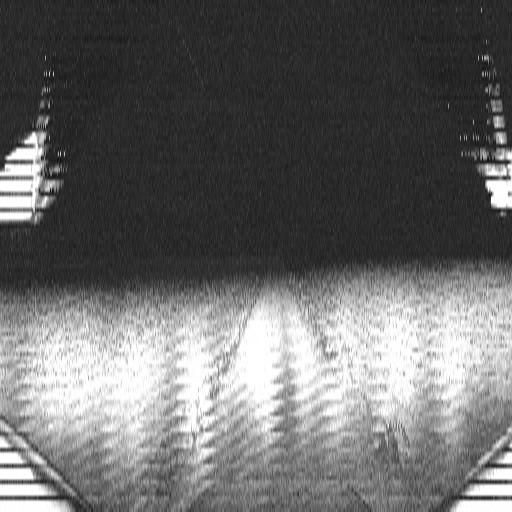}
\end{subfigure}
\begin{subfigure}{0.23\textwidth}
\centering
\includegraphics[width=\textwidth]{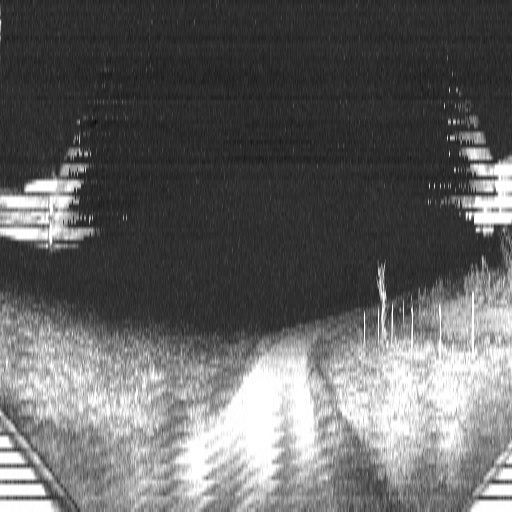}
\end{subfigure}

\begin{subfigure}{0.23\textwidth}
\centering
\includegraphics[width=\textwidth]{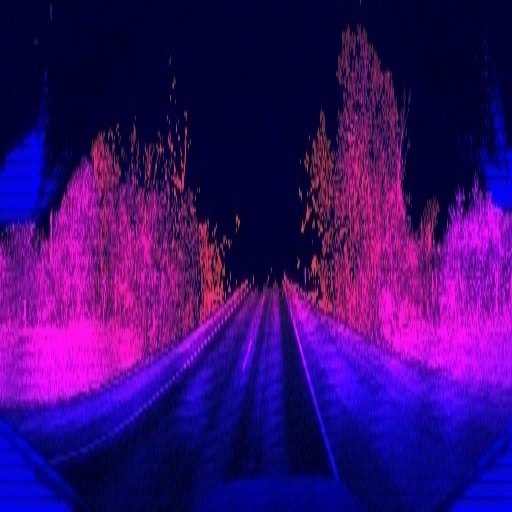}
\caption{Structured road}
\end{subfigure}
\begin{subfigure}{0.23\textwidth}
\centering
\includegraphics[width=\textwidth]{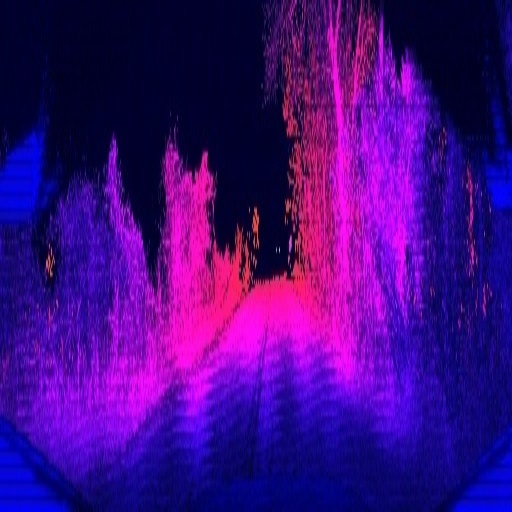}
\caption{Rural Single Lane Road}
\end{subfigure}
\begin{subfigure}{0.23\textwidth}
\centering
\includegraphics[width=\textwidth]{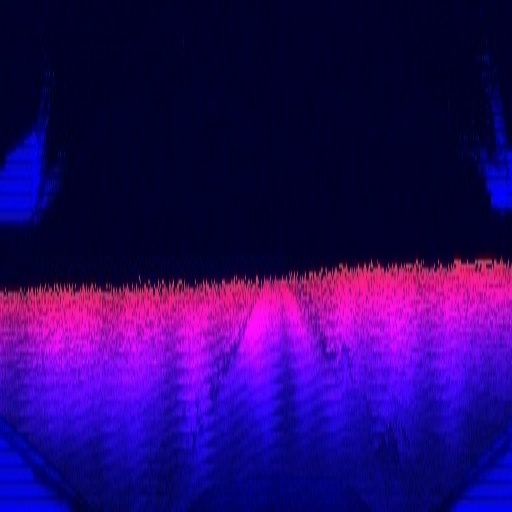}
\caption{Roads flanked by farm land}
\end{subfigure}
\begin{subfigure}{0.23\textwidth}
\centering
\includegraphics[width=\textwidth]{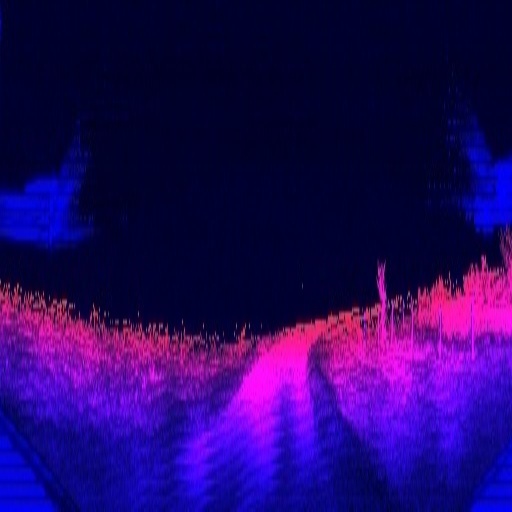}
\caption{Unpaved dirt based road}
\end{subfigure}
\caption{Samples images from the data-set; From top to bottom - \((i)\) Camera Image,\((ii)\) Single channel range image with intensity returns, and \((iii)\) Three channel range image with depth, intensity and reflectivity.}
\label{fig:rural_road_samples}
\end{figure*}

\subsection{Localization}

\begin{algorithm}
    
	\caption{Particle Filter}
	\begin{algorithmic}[1]
	    \For{number of particles}
	    \State Generate \(x_{t-1}\) using initial distribution
	    \EndFor
		\For {\(x_{t-1} \in X_{t-1}\)}
		\State Generate \(\overline{x}_{t}\) using motion model (Equation 4)
	    \EndFor
		\For {\(\overline{x}_{t} \in \overline{X}_{t-1}\)}
		\State Project point cloud P to map frame
		\For {\(p_i \in P\)}
	    \State Lookup distance to nearest edge \(d_{p_i}\)
	    \State \(w_t = w_t \times D(d_{p_i})\) 
		\EndFor
		\EndFor
		\If{steps \(>\) resampling count}
		\State Resample using weights \(W_t\)
		\EndIf
		\State \(x_t\) = Weighted average average of \(x_t \in X_{t}\)

	\end{algorithmic} 
\label{alg1}
\end{algorithm} 

The localization module combines information from Open Street Maps with road segmentation results and wheel odometry, through a variation of the Particle Filter. In context of the particle filter, a particle represents an estimate of the vehicle's pose and is defined as, \(x = [e,n,\theta]^T\), where \(e,n\) are map coordinates representing position, while \(\theta\) represents the orientation. The control input \(u_{t} = [de,dn,d\theta]^T\) represents the estimated change in pose between time steps \(t-1\) and \(t\). 

The sensor measurement \(z_t\) is a segmented point cloud, where each point \(p=[a,b,c]^T\), contains point coordinates \(a,b\) in the sensor frame of reference and a classification \(c \in [0,1]\) to identify road points. If the control inputs \(u_t\) and sensor measurements \(z_t\) between time \(1:t\) are, \(u_{1:t}\) and \(z_{1:t}\), the optimal pose is given as :
$$
x_t=\max_x p(x|u_{1:t},z_{1:t}) \eqno{(1)}
$$
The three step procedure is illustrated in Algorithm \ref{alg1}. First, a set of pose estimates \(X_{t-1}\) is generated based on a predefined distribution. The choice of this distribution depends on whether an initial pose is available or not. We sample from a uniform distribution, considering that initial pose is unknown. Next, we use wheel speed \(V\) and steering measurements \(\delta\) to compute \(u_t\) using a bicycle model as follows:
$$
 \omega = {V\times L}/\tan{\delta} , \space\theta = \int_0^t \omega.dt \eqno{(2)}
$$
$$
e = \int_0^t V\cos{\theta}.dt, n = \int_0^t V\sin{\theta}.dt \eqno{(3)}
$$
In the above equations, \(L\) is the wheel base and \(\omega\) is the angular velocity. The control input \(u_t\) along with initial estimates, \(X_{t-1}\) are used to generate a pose hypothesis,  \(\overline{X}_{t}\) by sampling over the probability distribution: 
$$
p(x_{t}|x_{t-1},u_{t}) \eqno{(4)}
$$

Next, an importance factor \(w_{t}\in W_t\) is assigned to each pose hypothesis \(\overline{x}_{t}\in \overline{X}_{t}\). The pose \(x_{t}\) is then obtained by sampling from \(\overline{X}_{t}\) based on the distribution represented by \(W_{t}\). To estimate the importance factor for a pose hypothesis \(\overline{x}_{t}\) we project the segmented point cloud onto \(\overline{x}_t\), and based on the distance \(d_{p_i}\) between the nearest road edge and the point \(p_i\) in the segmented cloud we estimate \(w_{t}\) as:
$$
w_{t}(\overline{x}_{t}) = \prod_{i=1}^{1=n}D(d_{p_i}) \eqno{(5)}
$$

where $D(.)$ represents the distance function and \(n\) is the number of points in the point cloud. We evaluate four distance functions - Gaussian with zero mean as used by \cite{ruchti2015localization}, a quadratic distance function (Equation 6), exponential decay (Equation 7) and a linear function proposed in Maplite\cite{ort2019maplite}. For road points (\(c=1\)) the quadratic and exponential decay functions \(D_{quadratic}\) and \(D_{exp.decay}\) are given as:
$$
D_{quadratic}(d_p) = \frac{1}{d_{p}^2 +1} \eqno{(6)}
$$
$$
D_{exp.decay}(d_p) = e^{-d_{p}\tau} \eqno{(7)}
$$
For \(c=0\), \(D_{p_{nonroad}}(d_p) = 1- D_{p_{road}}(d_p)\). The final pose estimate \(x_t\) is the weighted average over all \(x_t \in X_t\).

\section{Rural Road data-set}\label{data-set}

\subsection{Sensor Setup}
Our sensor setup consists of an RGB camera along with an Ouster OS1-128 LIDAR for collecting point cloud data. The LIDAR is run using  a \(180^\circ\) Field of View setting. To better capture the differences in material, color and texture we consider using a broader sensing spectrum. We therefore include intensity and  reflectivity of returns along with range measurements.

\subsection{Point Cloud Projection}
To project the point cloud to a perspective view image, each point is represented in terms of its spherical coordinates \(r,\theta,\phi\) where, 
$$
Range, \space r=\sqrt{x^2+y^2+z^2}, \eqno{(8)}
$$
$$
Inclination,\space\theta = \arccos{(z/r)}, \eqno{(9)}
$$
$$
Azimuth,\space\phi = atan2(y/x). \eqno{(10)}
$$

 \begin{figure*}[h!]
\centering
\begin{subfigure}{0.42\textwidth}
\centering
\includegraphics[width=\textwidth]{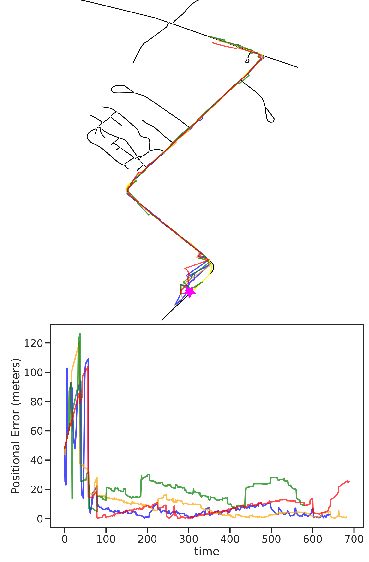}
\caption{Tracking}
\end{subfigure}
\begin{subfigure}{0.45\textwidth}
\centering
\includegraphics[width=\textwidth]{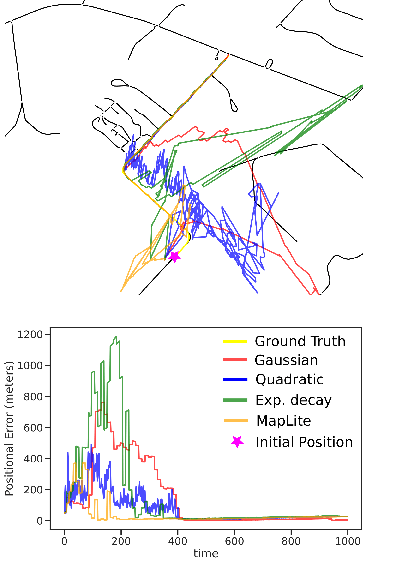}
\caption{Global Localization}
\end{subfigure}
\caption{Localization Performance on previously seen roads.}
\label{fig:seen_loc}
\end{figure*}
The sensor has a horizontal angular resolution of \(0.35^\circ\), the azimuth angle \(\phi\) of a point, thus maps it to a column of the 2D image. Likewise, the inclination \(\theta\) of a point, along with beam altitude angles map it to a row of the image. The final images thus have a $128\times 512$ resolution. We offer a single channel image containing intensity returns, as well as 3 channel images with intensity, reflectivity and range.

\subsection{Road Surface}
 The data-set includes samples of the following scenes:
\begin{itemize}
\item {\it Highways}: Representing well structured roads, with lane markings and other traffic signs.
\item {\it Rural single lane roads}: Characterized by vegetation on either sides.
\item {\it Farm Lands}: Roads that are flanked by farm land on both sides.
\par
\end{itemize}
Samples from the data-set are shown in Figure \ref{fig:rural_road_samples}. It is worth noting that 80\% of our data is from rural environments, greater than any of the previously existing data-sets. 
\subsection{Annotation}
The range images are manually annotated and saved as VOC style annotations. These annotations are used to generate segmentation masks. A total of 2818 images were annotated, of which 2600 are used for training and 218 are used for validation.
 \par

 \begin{figure*}[h!]
\centering
\begin{subfigure}{0.45\textwidth}
\centering
\includegraphics[width=\textwidth]{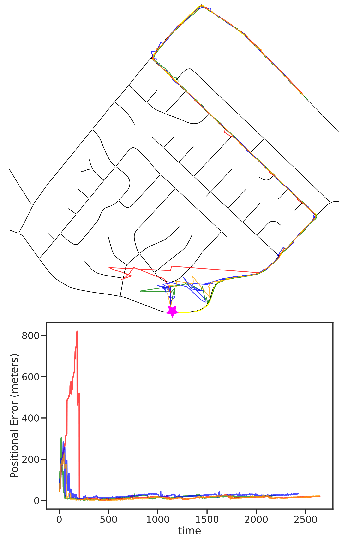}
\caption{Tracking}
\end{subfigure}
\begin{subfigure}{0.435\textwidth}
\centering
\includegraphics[width=\textwidth]{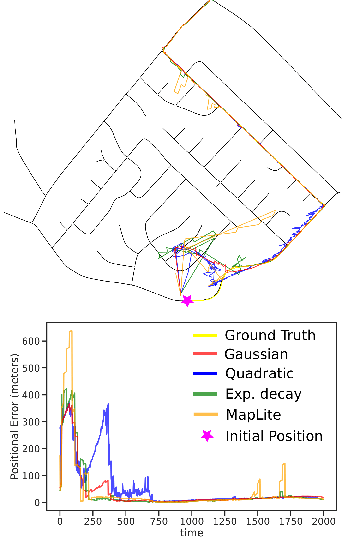}
\caption{Global Localization}
\end{subfigure}
\caption{Localization performance on previously unseen roads.}
\label{fig:unseen_loc}
\end{figure*}
\section{Experimentation and Results}\label{train}
\subsection{Training}
We evaluate our data-set on three state of the art image segmentation models DeepLabv3 \cite{chen2017rethinking}, UNet \cite{ronneberger2015u}, and FCN \cite{Lin_2017_CVPR}. Like most image segmentation models, these use an encoder-decoder architecture. ResNet 50 backbone is used for each of the models, with a batch size of 4  images. Adam optimization  with a learning rate of \(1e-3\) was used and layer weights are initialized using Xavier initialization. To account for the smaller percentage of road points in most images, we use weighted cross entropy loss:\par
$$
loss = W.y_{t}(-\log(\sigma(y_{p}))) + (1 - y_{t})(-\log(1 - \sigma(y_{p}))) \eqno{(7)}
$$
where \(y_{t}\), \(y_{p}\) represent the ground truth and model prediction respectively, \(\sigma\) represents the sigmoid function and \(W\) represents the point weights based on class. Using this scheme road points are give three times more weight than non-road points.\par
Models are each trained for 200 epochs on a Nvidia RTX 2070 GPU. We use four metrics for model performance evaluation: mean IoU, Precision, Recall and F1 score. 

\par

    
     

\subsection{Evaluation}
The model performance on validation data is shown in Table \ref{table:mchanne;}, where models with subscript \(m\) denote their multi-channel variants. It can be seen that these 2D segmentation models are not only able to segment road surfaces, but they do so with a remarkably high accuracy, e.g. the mean IoU for DeepLabv3 is 92.5\% which is on par  with best performing algorithms on the KITTI Road Leader-board. It is seen that DeepLabv3 outperforms the other models for both single channel and multi channel images.

\begin{table}[b]
    \caption{Model performance on validation set}
    \label{table:mchanne;}
    \centering
    
    \begin{tabular}{ p{1.8cm} p{1.2cm} p{1cm} p{1cm} p{1cm}   }
         \hline
     Model & Precision & Recall & F1 & mIoU\\
     
     \hline
     \(UNet\)   &\textbf{93.57}    &90.16&   91.64 & 81.91\\
     \(UNet_m\)   & 93.07    &91.24&   91.83 & 82.2\\
     
     \(FPN\) &   89.4  & 85.01   & 87.06 & 77.28\\
     \(FPN_m\) &  88.85  & 85.33   &86.82 & 76.88\\
     
     \(Deeplabv3\) &92.71 & \textbf{96.22} &  \textbf{94.34} & \textbf{92.49}\\
     \(Deeplabv3_m\) &92.51 & 93.75&  93.09 & 83.79\\
    \hline
    \end{tabular}
    \newline

\end{table}

\subsection{Localization Performance}
The DeepLabv3 model is used in conjunction with the localization module, to evaluate localization performance for tracking and global localization. A GNSS receiver with 2.0 meter horizontal sensing accuracy is used for ground-truth measurements of position. The motion model in Equation 4, is represented by a Gaussian with standard deviation of \(0.1\) meters for position and \(3^{\circ}\) for orientation. To reduce runtimes, we down sample the point cloud using a voxel grid, with voxels of size \(2\times 2\) meters, and pre-compute distance to nearest edge for all points on the map. Particles are sampled according to \(W_{t}\)  every 20 time steps.\par
For tracking tests, 10,000 particles within a 200m radius of the initial pose estimate are used, while global localization uses 100,000 particles uniformly distributed over a \(2\) sq.km area. Each of the four distance functions is evaluated and localization performance is compared. Global localization is used to evaluate rate of convergence and tracking is used to evaluate positional error after convergence. The results are reported  in Table \ref{table:loc_res}. \par

As a first test, we use a route which has been seen by the road segmentation model, as training data was collected on this route. The positional error as a function of time as well as a map overlaid with the weighted average pose estimate is shown in Figure \ref{fig:seen_loc}.
\begin{table}[h]
    \caption{Localization performance for different weighting functions}
    \label{table:loc_res}
    \centering
    
    \begin{tabular}{ p{1.3cm} p{0.8cm} p{0.8cm} p{1.cm} p{1.5cm}   }
         \hline
    Distance Function & \multicolumn{2}{c}{ Mean Error (meters)} & \multicolumn{2}{c}{Global Convergence time (steps) }\\
     & Test-1  & Test-2 & Test-1  & Test-2\\

     \hline
     Gaussian    &   6.7 & 9.4&   400&   \textbf{350} \\
     Quadratic  & \textbf{4.79}    & 15.65&   400 &   650\\
     
     Exp. decay &   19.1  & 4.95   & \textbf{220} & \textbf{350}   \\
     Maplite &  6.71  & \textbf{4.07}   & \textbf{220} &   600\\
     
    \hline
    \end{tabular}
    \newline

\end{table}
For tracking, it is seen that all functions converge after 100 steps, with  Quadratic and Gaussian functions having the lowest mean error after convergence of 4.79 and 6.7 meters respectively . For global localization, it is seen that all the functions converge after around 400 steps, of which Maplite and exponential decay converge the fastest. 

A previously unseen route is used for the second test. The particle filter is initialized with same parameters as the first case and the results are shown in Figure \ref{fig:unseen_loc}. We see that the Maplite function has the smallest error of 4.07 meters for tracking, and once again  Exponential decay and Gaussian functions have the fastest convergence. It can also be seen that the Quadratic function requires a significantly longer time to converge as compared to the first test, which in turn increases the mean error. This suggests that the localization convergence is most impacted by performance of the road segmentation module. From these tests, it is observed that Gaussian and Exponential Decay  functions are most preferable for fast convergence over a large search space. In comparison, Maplite and Quadratic functions produce the lowest mean error once the search space is narrowed down.

\section{Conclusion}
A novel rural road data-set is presented, which when used to train existing image segmentation models, is able to achieve performance levels comparable to models designed specifically for point cloud segmentation. The proposed model is used to implement a LIDAR based, GPS denied localization algorithm for both tracking and global localization. Experimental results demonstrate that the algorithm is successfully able to estimate pose, with a mean error of \(4\) meters for tracking, and \(6.5\) meters for global localization. A combination of Gaussian or Exponential decay for fast initial convergence followed by the Maplite function for low tracking errors can further enhance the performance of the algorithm.





\section*{ACKNOWLEDGMENT}

\noindent Support for this research was provided in part by a grant from the U.S. Department of Transportation, University Transportation Centers Program to the Safety through Disruption University Transportation Center (451453-19C36). \\

 \noindent {\textbf{Disclaimer:}} The contents of this paper reflect the views of the authors, who are responsible for the facts and the accuracy of the information presented herein. This document is disseminated in the interest of information exchange. The report is funded, partially or entirely, by a grant from the U.S. Department of Transportation’s University Transportation Centers Program. However, the U.S. Government assumes no liability for the contents or use thereof.


\bibliographystyle{IEEEtran}
\bibliography{IEEEabrv,root}
\addtolength{\textheight}{-12cm}   


\end{document}